\documentclass[conference,compsoc]{IEEEtran}
\usepackage{placeins}
\usepackage{amsmath,graphicx}
\usepackage{multirow}
\usepackage[font=small,skip=0pt]{caption}

\ifCLASSOPTIONcompsoc
  \usepackage[nocompress]{cite}
\else
  \usepackage{cite}
\fi

\begin{document}
		\nocite{*}
		
\title{Simplified Long Short-term Memory \\ Recurrent Neural Networks: part II}

\author{\IEEEauthorblockN{Atra Akandeh and Fathi M. Salem}
\IEEEauthorblockA{Circuits, Systems, and Neural Networks (CSANN) Laboratory \\
Computer Science and Engineering , Electrical and Computer Engineering \\
Michigan State University\\
East Lansing, Michigan 48864-1226\\
akandeha@msu.edu; salemf@msu.edu }
}

\maketitle

\begin{abstract}
This is part II of three-part work. Here, we present a second set of inter-related five variants of simplified Long Short-term Memory (LSTM) recurrent neural networks by further reducing
adaptive parameters. Two of these models have been introduced in part I of this work. We evaluate and verify our model variants on the benchmark MNIST dataset and assert that these models are comparable to the base LSTM model while use progressively less number of parameters. Moreover, we observe that in case of using the ReLU activation, the test accuracy performance of the standard LSTM will drop after a number of epochs when learning parameter become larger. However all of the new model variants sustain their performance.
\end{abstract}


\section{Introduction}
In contrast to simple Recurrent Neural Networks (RNNs), Gated RNNs are more powerful in performance when considering sequence-to-sequence relationships [1-8]. The simple RNN model can be mathematically expressed as:
\begin{equation}
	\begin{split}
		& h_t = \sigma(W_{hx} x_t + W_{hh} h_{t-1} + b_h) \\
		& y_t = W_{hy} h_t + b_y
	\end{split}
\end{equation}

where $W_{hx}$, $W_{hh}$, $b_{h}$, $W_{hy}$ and $b_{y}$ are adaptive set of weights and $\sigma$ is a nonlinear function. In the base LSTM model, the usual activation function has been equivalently morphed into a more complicated activation function, so that the hidden units enable the back propagated through time (BBTT) gradients technique to function properly \cite{lstm}. The base LSTM uses memory cells in the base network with incorporated three gating mechanisms to properly process the sequence data. To do so, the base LSTM models  introduce new sets of parameter in the gating signals and hence more computational cost and slow training speed. The base (standard) LSTM model is expressed as 
\begin{equation}
	\begin{split}
		& i_t = \sigma_{in}(W_i x_t + U_i h_{t-1} + b_i) \\
		& f_t = \sigma_{in}(W_f x_t + U_f h_{t-1} + b_f) \\
		& o_t = \sigma_{in}(W_o x_t + U_o h_{t-1} + b_o) \\
		& \tilde{c_t} = \sigma(W_c x_t + U_c h_{t-1} + b_c) \\
		& c_t= f_t \odot c_{t-1} + i_t \odot \tilde{c_t}\\
		& h_t = o_t \odot \sigma(c_t) 
	\end{split}
\end{equation}
Note that the first three equations represent the gating signals, while the remaining three equations express the memory-cell networks. In our previous work we have introduced 5 reduced variants of the base LSTM, referred to as 
$\textbf{LSTM1, ~LSTM2, ~ LSTM3,}$ and $\textbf{LSTM5}$, respectively. We will use the same numbering designation, and continue the numbering system in our publications to distinguish among the model variants.

\section{Parameter-reduced variants of the LSTM base model}
A unique set of five variants of the base LSTM model are introduced and evaluated here. Two of them have been presented in part I and are used in this comparative evaluation. To facilitate identification, we designate odd-numbered variants to include biases and even-numbered variants to have no biases in the gating signals.

\subsection{LSTM4}
LSTM4 has been introduced in part I. We have removed the input signal and the bias from the gating equations. Furthermore, the matrices $U_i$, $U_f$ and $U_o$ are replaced with the corresponding vectors  $u_i$, $u_f$ and $u_o$ respectively, in order to render a point-wise multiplication. Specifically, one obtains

\begin{equation}
	\begin{split}
		& i_t = \sigma_{in}(u_i  \odot  h_{t-1}) \\
		& f_t = \sigma_{in}(u_f \odot  h_{t-1}) \\
           & o_t = \sigma_{in}(u_o \odot  h_{t-1}) \\
		& \tilde{c_t} = \sigma(W_c x_t + U_c h_{t-1} + b_c) \\
		& c_t= f_t \odot c_{t-1} + i_t \odot \tilde{c_t}\\
		& h_t = o_t \odot \sigma(c_t) 
	\end{split}
\end{equation}

\subsection{LSTM5}
LSTM5 was also presented in part I and it is similar to LSTM4 except that the biases are retained in the three gating signal equations.
\begin{equation}
	\begin{split}
		& i_t = \sigma_{in}(u_i  \odot h_{t-1} + b_i) \\
		& f_t = \sigma_{in}(u_f \odot h_{t-1} + b_f) \\
		& o_t = \sigma_{in}(u_o \odot h_{t-1} + b_o) \\
		& \tilde{c_t} = \sigma(W_c x_t + U_c h_{t-1} + b_c) \\
		& c_t= f_t \odot c_{t-1} + i_t \odot \tilde{c_t}\\
		& h_t = o_t \odot \sigma(c_t) 
	\end{split} 
\end{equation}

\subsection{LSTM4a}
In LSTM4a, a fixed real number with absolute value less than 1 has been set for the forget gate in order to preserve bounded-input-bounded-output (BIBO) stability, \cite{salem2016_basic}.  Meanwhile, the output gate is set to 1 (which in practice eliminates this gate altogether).
\begin{equation}
	\begin{split}
		& i_t = \sigma_{in}(u_i \odot h_{t-1}) \\
		& f_t = 0.96 \\
		& o_t = 1.0 \\
		& \tilde{c_t} = \sigma(W_c x_t + U_c h_{t-1} + b_c) \\
		& c_t= f_t \odot c_{t-1} + i_t \odot \tilde{c_t}\\
		& h_t = o_t \odot \sigma(c_t) 
	\end{split}
\end{equation}

\subsection{LSTM5a}
LSTM5a is similar to LSTM4a but the bias term in the input gate equation is preserved.
\begin{equation}
	\begin{split}
		& i_t = \sigma_{in}(u_i \odot h_{t-1} + b_i) \\
		& f_t = 0.96 \\
		& o_t = 1.0 \\
		& \tilde{c_t} = \sigma(W_c x_t + U_c h_{t-1} + b_c) \\
		& c_t= f_t \odot c_{t-1} + i_t \odot \tilde{c_t}\\
		& h_t = o_t \odot \sigma(c_t) 
	\end{split}
\end{equation}

\subsection{LSTM6}
Finally, this is the most aggressive parameter reduction. Here all gating equations replaced replaced by appropriate constant. For BIBO stability, we found that the forget gate must be set $f_t = 0.59$ or below. The other two gates are set to 1 each (which practically eliminate them to the purpose of computational efficiency). In fact, this model variant now becomes equivalent to the so-called basic RNN model reported in \cite{salem2016_basic}.
\begin{equation}
	\begin{split}
		& i_t = 1.0 \\
		& f_t = 0.59 \\
		& o_t = 1.0 \\
		& \tilde{c_t} = \sigma(W_c x_t + U_c h_{t-1} + b_c) \\
		& c_t= f_t \odot c_{t-1} + i_t \odot \tilde{c_t}\\
		& h_t = o_t \odot \sigma(c_t) 
	\end{split}
\end{equation}

\begin{table}
	\caption{variants specifications.}
	\centering
	\begin{tabular}{| c | c | c |} 
		\hline
		variants & \# of parameters & times(s) per epoch \\ 
		\hline
		LSTM & 52610 & 30 \\
		\hline
		LSTM4 & 14210 & 23 \\
		\hline
		LSTM5 & 14510 & 24 \\
		\hline
		LSTM4a & 14010 & 16 \\
		\hline
		LSTM5a & 14110 & 17 \\
		\hline
		LSTM6 & 13910 & 12 \\
		\hline
	\end{tabular}
	\label{vs}
\end{table}

Table~\ref{vs} provides the number of total parameters as well as the training times per epoch corresponding to each variant. As shown, the number of parameters and time per epoch have been progressively decreased to less than half of the corresponding values for the base (standard) LSTM model.
\FloatBarrier
\section{Experiments and Discussion}
To perform equitable comparison among all variants, similar condition have been adopted using the Keras Library environment \cite{keras_rowwise}. The model specification are depicted in Table~\ref{tab:a}. We have trained the models using the row-wise fashion of the MNIST dataset. Each image is $28 \times 28$. Hence, sequence duration is 28 and input dimension is also 28. There are three case studies involving distinct activation: $tanh$, $sigmoid$ and $relu$ of first layer. For each case, we have tuned the learning parameter $\eta$ over three values.
\begin{table}[!htb]
	\caption{Network specifications.}
	\centering
	\begin{tabular}{| c | c |} 
		\hline
		Input dimension & $28 \times 28 $ \\ 
		\hline
		Number of hidden units & 100 \\
		\hline
		Non-linear function & tanh, sigmoid, tanh \\
		\hline
		Output dimension & 10 \\
		\hline
		Non-linear function & softmax \\
		\hline
		Number of epochs / Batch size & $100 / 32$ \\
		\hline
		Optimizer / Loss function & RMprop / categorical cross-entropy\\
		\hline
	\end{tabular}
	\label{tab:a}
\end{table}

\subsection{The tanh activation}
The activation function is $tanh$. To improve performance, learning parameter over only three different values is evaluated. In training, initially there is a gap among different model variants, however, they all catch up with the base (standard) LSTM quickly within the 100 epochs. Test accuracy of base LSTM model is at 99\% and test accuracy of other variants after parameter tuning is around 98\%. As one increases $\eta$, more fluctuation is observed. However, the models still sustain their performance levels. One advantage of the model variant LSTM6 is that it rises faster in comparison to other variants. However, $\eta = 0.002$ causes it to decrease, indicating that this $\eta$ value is relatively large. In all cases, LSTM4a and LSTM5a performance curves overlap. 

\begin{figure}[!htb]
	\vspace{-4mm}
	\centering
	\setlength{\belowcaptionskip}{-20pt}
	\includegraphics[trim={0 0 0 0},clip,scale=0.42]{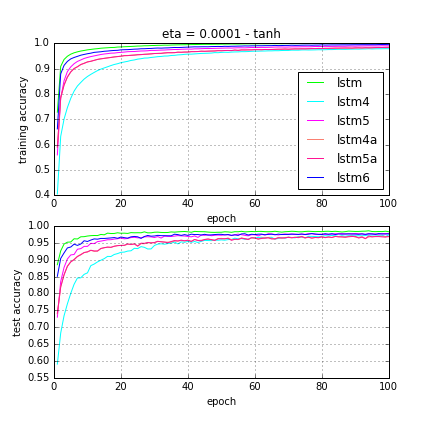}
	\caption{Training \& Test accuracy, $\sigma=tanh , \eta=1\mathrm{e}{-4}$}
	\label{fig:fig1}
\end{figure}

\begin{figure}[!htb]
	\centering
	\setlength{\belowcaptionskip}{-5pt}
	\includegraphics[trim={0 0 0 0},clip,scale=0.42]{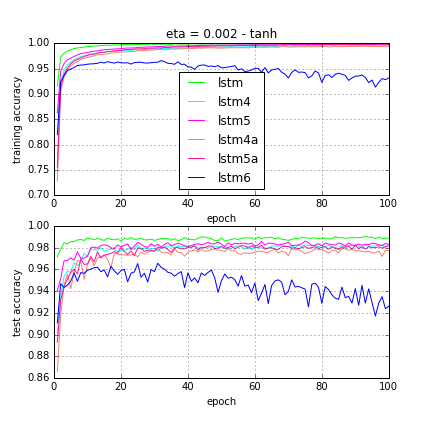}
	\caption{Training \& Test accuracy, $\sigma=tanh , \eta=2\mathrm{e}{-3}$}
	\label{fig:fig3}
\end{figure}

The best results obtained among over the 100 epochs training duration is summarized in table~\ref{tanh}. For each model variant, the best results are shown in bold font.
\begin{table}[!htb]
	\caption{Best results obtained by the tanh activation.}
	\centering
	\begin{tabular}{ |c|c|c|c|c| }
		\cline{3-5}
		\multicolumn{1}{c}{}
		& & $\eta = 1\mathrm{e}{-4}$ & $\eta =1\mathrm{e}{-3}$ & $\eta = 2\mathrm{e}{-3}$ \\
		\hline
		\multirow{2}{*}{LSTM} & train &  0.9995 &  1.0000 &  0.9994 \\ 
		& test &  0.9853 &  0.9909 &  \textbf{0.9903} \\ 
		\hline
		\multirow{2}{*}{LSTM4} & train &  0.9785 &  0.9975 &  0.9958 \\ 
		& test &  0.9734 &  \textbf{0.9853} &  0.9834 \\ 
		\hline
		\multirow{2}{*}{LSTM5} & train &  0.9898 &  0.9985 &  0.9983 \\ 
		& test &  0.9774 &  0.9835 &  \textbf{0.9859} \\ 
		\hline
		\multirow{2}{*}{LSTM4a} & train &  0.9835 &  0.9957 &  0.9944 \\ 
		& test &  0.9698 &  \textbf{0.9803} &  0.9792 \\ 
		\hline\
		\multirow{2}{*}{LSTM5a} & train &  0.9836 &  0.9977 &  0.998 \\ 
		& test &  0.9700 &  0.9820 &  \textbf{0.9821} \\ 
		\hline\
		\multirow{2}{*}{LSTM6} & train &  0.9948 &  0.9879 &  0.9657 \\ 
		& test &  0.9771 &  \textbf{0.9792} &  0.9656 \\ 
		\hline
	\end{tabular}
	\label{tanh}
\end{table}

\subsection{The (logistic) sigmoid activation}
For these cases the similar trend is observed. The only difference is that the sigmoid activation with $\eta=1\mathrm{e}{-4}$ slowly progresses towards its maximal performance. After learning parameter tuning, the typical test accuracy of the base LSTM model is 99\%, test accuracy of variants LSTM4 and LSTM5 is about 98\%, and, test accuracy of variants LSTM4a and LSTM5a is 97\%. As it is shown in the table and associated plots, $\eta=1e-4$ seemingly relatively small when using the $sigmoid$ activation. As evidence, LSTM6 attains only a training and test accuracies of about 74\% after 100 epochs, while it attains an accuracy of about 97\% when $\eta$ is increased 10-fold. In our variants, models using $tanh$ saturate quickly; however, models using $sigmoid$ rise steady from beginning to the last epoch. In these cases, again the LSTM4a and LSTM5a performances overlap.
\begin{figure}[!htb]
	\vspace{-4mm}
	\centering
	\setlength{\belowcaptionskip}{-20pt}
	\includegraphics[trim={0 0 0 0},clip,scale=0.42]{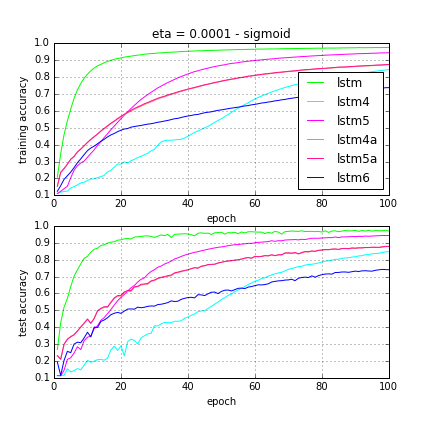}
	\caption{Training \& Test accuracy, $\sigma=sigmoid , \eta=1\mathrm{e}{-4}$}
	\label{fig:fig4}
\end{figure}

\begin{figure}[!htb]
	\centering
	\includegraphics[trim={0 0 0 0},clip,scale=0.42]{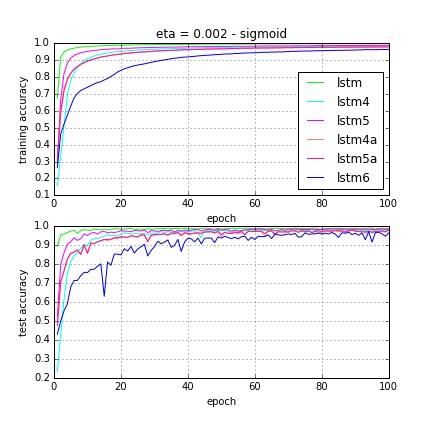}
	\caption{Training \& Test accuracy, $\sigma=sigmoid , \eta=2\mathrm{e}{-3}$}
	\label{fig:fig6}
\end{figure}

\begin{table}[!htb]
		\caption{Best results obtained by sigmoid model.}
		\centering
	\begin{tabular}{ |c|c|c|c|c| }
		\cline{3-5}
		\multicolumn{1}{c}{}
		& & $\eta = 1\mathrm{e}{-4}$ & $\eta =1\mathrm{e}{-3}$ & $\eta = 2\mathrm{e}{-3}$ \\
		\hline
		\multirow{2}{*}{LSTM} & train &  0.9751 &  0.9972 &  0.9978 \\ 
		& test &  0.9739 &  0.9880 &  \textbf{0.9886} \\ 
		\hline
		\multirow{2}{*}{LSTM4} & train &  0.8439 &  0.9793 &  0.9839 \\ 
		& test &  0.8466 &  0.9781 &  \textbf{0.9822} \\ 
		\hline
		\multirow{2}{*}{LSTM5} & train &  0.9438 &  0.9849 &  0.9879 \\ 
		& test &  0.9431 &  0.9829 &  \textbf{0.9844} \\ 
		\hline
		\multirow{2}{*}{LSTM4a} & train &  0.8728 &  0.9726 &  0.9778 \\ 
		& test &  0.8770 &  0.9720 &  \textbf{0.9768} \\ 
		\hline
		\multirow{2}{*}{LSTM5a} & train &  0.8742 &  0.9725 &  0.9789 \\ 
		& test &  0.8788 &  0.9707 &  \textbf{0.9783} \\ 
		\hline
		\multirow{2}{*}{LSTM6} & train &  0.7373 &  0.9495 &  0.9636 \\ 
		& test &  0.7423 &  0.9513 &  \textbf{0.9700} \\ 
		\hline
	
	\end{tabular}
	\label{sigmoid}
\end{table}

\subsection{The relu activation}
For the $relu$ activation cases, we tune the performance over three learning rate parameters for comparison. All models perform well with $\eta=1e-4$.  In this case, LSTM4a and LSTM5a do not overlap. For $\eta=1e-3$, the base LSTM does not sustain its performance and drastically drops. One may need to leverage early stopping strategy to avoid this problem. In this case study, the LSTM model begin to fall around epoch=50. LSTM6 also drops in performance. Model LSTM4, LSTM5, LSTM4a and LSTM5a show sustained accuracy performance as for $\eta=1e-4$. For $\eta=2e-3$ model LSTM4 and LSTM5 are still sustaining their performances. 

\begin{figure}[!htb]
	\vspace{-4mm}
	\centering
	\includegraphics[trim={0 0 0 0},clip,scale=0.42]{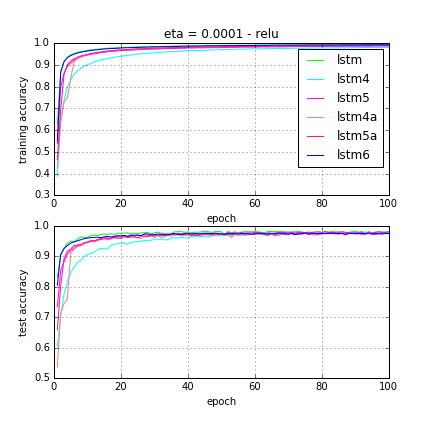}
	\caption{Training \& Test accuracy, $\sigma=relu , \eta=1\mathrm{e}{-4}$}
	\label{fig:fig7}
\end{figure}

\begin{figure}[!htb]
	\vspace{-8mm}
	\centering
	\includegraphics[trim={0 0 0 0},clip,scale=0.42]{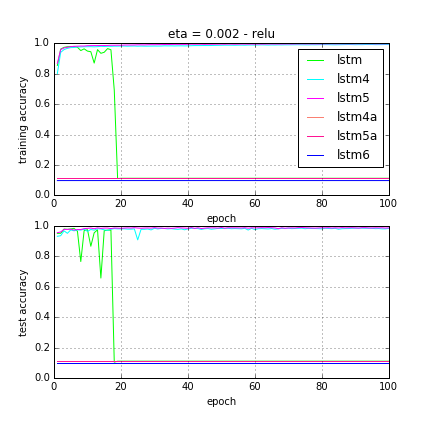}
	\caption{Training \& Test accuracy, $\sigma=relu , \eta=2\mathrm{e}{-3}$}
	\label{fig:fig9}
\end{figure}

The best results obtained among all the epochs have been shown in Table~\ref{relu}.
\begin{table}[!htb]
	\caption{Best results obtained by relu model.}
	\centering
	\begin{tabular}{ |c|c|c|c|c| }
		\cline{3-5}
		\multicolumn{1}{c}{}
		& & $\eta = 1\mathrm{e}{-4}$ & $\eta =1\mathrm{e}{-3}$ & $\eta = 2\mathrm{e}{-3}$ \\
		\hline
		\multirow{2}{*}{LSTM} & train &  0.9932 &  0.9829 &  0.9787 \\ 
		& test &  0.9824 &  \textbf{0.9843} &  0.9833 \\ 
		\hline
		\multirow{2}{*}{LSTM4} & train &  0.9808 &  0.9916 &  0.9918 \\ 
		& test &  0.9796 &  \textbf{0.9857} &  0.9847 \\ 
		\hline
		\multirow{2}{*}{LSTM5} & train &  0.987 &  0.9962 &  0.9964 \\ 
		& test &  0.9807 &  0.9885 &  \textbf{0.9892} \\ 
		\hline
		\multirow{2}{*}{LSTM4a} & train &  0.9906 &  0.9949 &  0.1124 \\ 
		& test &  0.9775 &  \textbf{0.9878} &  0.1135 \\ 
		\hline
		\multirow{2}{*}{LSTM5a} & train &  0.9904 &  0.996 &  0.1124 \\ 
		& test &  0.9769 &  \textbf{0.9856} &  0.1135 \\ 
		\hline
		\multirow{2}{*}{LSTM6} & train &  0.9935 &  0.9719 &  0.09737 \\ 
		& test &  \textbf{0.9761} &  0.9720 &  0.0982 \\ 
		\hline
	\end{tabular}
	\label{relu}
\end{table}

\section{Conclusion}
We have aimed at reducing the computational cost and increasing execution times by presenting new variants of the (standard) base LSTM model.  LSTM4 and LSTM 5, which were introduced earlier,  use pointwise multiplication between the states and their corresponding weights. Model LSTM5, as all other odd-numbered model variants, retains the bias term in the gating signals. Model LSTM4a and Model LSTM5a are similar to model LSTM4 and LSTM5 respectively. The only difference is that the forget and the output gate are set at appropriate constants.  In model LSTM6, which is actually a basic recurrent neural network, all gating equations have been replaced by a constants! It has been demonstrated that all new variants are relatively comparable to the base (standard) LSTM in this initially case study. Using appropriate learning rate $\eta$ values, LSTM5, then LSTM4a \& LSTM5a, LSTM4 and LSTM6, respectively, have progressively close performance to the base LSTM. Finally we can conclude that any of introduced model variants, with hyper-parameter tuning, can be used to train a dataset with markedly less computational effort.

\section*{Acknowledgment}
This work was supported in part by the National Science Foundation under grant No. ECCS-1549517.

\small{
	\bibliographystyle{ieee}
	\bibliography{egbib}
}

\end{document}